\documentclass[11pt,a4paper]{article}
\usepackage[hyperref]{acl2021}
\usepackage{times}
\usepackage{latexsym}

\definecolor{lightyellow}{HTML}{eb9634}

\definecolor{lightblue}{rgb}{.90,.92,1}
\definecolor{lichen}{rgb}{.91,.95,0.83}

\usepackage{microtype}

\usepackage{color}
\usepackage{booktabs}
\usepackage{multicol}
\usepackage{graphicx}
\usepackage{xcolor}
\usepackage{caption}
\usepackage{subcaption}
\usepackage{wrapfig}
\usepackage{colortbl}
\usepackage{bm}
\usepackage{amssymb}
\usepackage{soul}
\usepackage{tabularx}
\usepackage{float}
\usepackage[htt]{hyphenat}
\usepackage{tabu}
\usepackage{amsmath,bm}
\usepackage{array,multirow,colortbl}
\usepackage{todonotes}
\usepackage{multirow}

\newenvironment{packed_enumerate}{
\begin{enumerate}
  \setlength{\itemsep}{1pt}
  \setlength{\parskip}{0pt}
  \setlength{\parsep}{0pt}
}{\end{enumerate}}

\aclfinalcopy % Uncomment this line for the final submission
\def\aclpaperid{4260} %  Enter the acl Paper ID here

\newcolumntype{Y}{>{\centering\arraybackslash}X}

\title{Elaborative Simplification: Content Addition and Explanation Generation in Text Simplification}

\author{Neha Srikanth \\
  Department of Computer Science \\
  The University of Texas at Austin \\
  \texttt{nehasrik@utexas.edu} \\\And
  Junyi Jessy Li \\
  Department of Linguistics \\
  The University of Texas at Austin \\
  \texttt{jessy@austin.utexas.edu} \\
 }

\date{}

\begin{document}
\maketitle
\begin{abstract}

Much of modern-day text simplification research focuses on sentence-level simplification, transforming original, more complex sentences into simplified versions. However, \emph{adding} content can often be useful when difficult concepts and reasoning need to be explained.
In this work, we present the first data-driven study of content addition in text simplification, which we call \emph{elaborative simplification}. We introduce a new annotated dataset of 1.3K instances of elaborative simplification in the Newsela corpus, and analyze how entities, ideas, and concepts are elaborated through the lens of contextual specificity. We establish baselines for elaboration generation using large-scale pre-trained language models, and demonstrate that considering contextual specificity during generation can improve performance. Our results illustrate the complexities of elaborative simplification, suggesting many interesting directions for future work.

\end{abstract}

\section{Introduction}

Text simplification aims to help audiences read and understand a piece of text through lexical, syntactic, and discourse modifications, while remaining faithful to its central idea and meaning~\cite{siddharthan2014survey}. It remains an important task, improving text accessibility for children~\cite{de2010text,kajiwara2013selecting}, language learners~\cite{yano1994effects,petersen2007text,pellow2014open,paetzold2016lexical}, and those with language impairments~\cite{carroll1998practical,rello2013frequent}. Text simplification can also be a useful pre-processing step for other NLP tasks such as machine translation~\cite{chen2012simplification,vstajner2016can} and summarization~\cite{vanderwende2007beyond,silveira2012enhancing}.

With the introduction of large, parallel corpora~\cite{zhu2010monolingual,woodsend2011learning,coster2011simple,xu2015problems}, text simplification research has rapidly advanced in recent years, especially in sentence simplification~\cite{alva2020data}. 
However, document simplification involves rich linguistic phenomena that cannot be easily characterized by sentence-level transformations of text, e.g., the omission and addition of content~\cite{petersen2007text,siddharthan2014survey}.

\begin{table}
{\fontfamily{cmr}\selectfont
    \centering
    \small
    \renewcommand{\arraystretch}{1.25}% Tighter
    \begin{tabular}{c} 
    \toprule
    \textbf{Original Text} 
    \\
    \arrayrulecolor[rgb]{0.753,0.753,0.753}\hline
    \multicolumn{1}{p{7.4cm}}{
    Results, she said, ``could help the team better understand ancient Egyptian health" and, correspondingly, modern-day health. For instance, some mummies still have arteries in their mummified remains, Miller-Thomas said. And, sometimes, scientists can tell if those arteries had hardened.
    } 
    \\
    \arrayrulecolor[rgb]{0,0,0} \hline
    \rule{0pt}{1.17\normalbaselineskip}
    \textbf{Simplified Text} 
    \\
    \arrayrulecolor[rgb]{0.753,0.753,0.753}\hline
    \multicolumn{1}{p{7.4cm}}{The scans could help the team understand about ancient Egyptians' health. For example, some mummies still have arteries. \sethlcolor{lichen}\hl{\textbf{An artery is a tube that moves blood through the body.}} \sethlcolor{lightblue}\hl{\textbf{The artery could show if the person had been healthy or not.}}} 
    \\
    \arrayrulecolor{black}
    \bottomrule
    \end{tabular}
}
\captionof{figure}{Elaborative simplification with two elaborations of varying contextual specificity.}
\label{tab:artery-example}
\vspace{-0.5mm}
\end{table}

This paper presents the first data-driven, dedicated study of \emph{elaborative simplification}, which involves inserting elaborations in the form of definitions, explanations or clarifications to improve readability by providing readers with necessary additional context. Effective elaborations must provide background in a \textit{contextual} manner, adding relevant information to the surrounding text. 

Figure \ref{tab:artery-example} shows an example. The original text snippet explains that scientists study mummy arteries to see whether they are hardened. In the corresponding simplified text, we see two elaborations inserted -- one, in green, simply defines an artery, and the second, in blue, states the implication of hardened arteries. The content of both elaborations is semantically absent from the original text.

Our goal is to provide resources and directions toward understanding and generating naturally occurring elaborations.
% \footnote{A related task is deciding \emph{when} to provide such elaboration; we leave this for future work as discussed in Section~\ref{sec:scope}.}
We present an annotated dataset of 1.3K instances of elaborative simplification in the Newsela corpus ~\cite{xu2015problems}. 
We automatically identify candidate elaborations from simplified documents, and have human annotators verify candidates. We find that many elaborations require multi-hop reasoning, inference, commonsense reasoning, and relevant information retrieval, making it an interesting testbed for a bevy of related tasks.

The previous example highlights two elaborations on opposite ends of the spectrum -- the first requires little context, while the second is highly contextualized, drawing a conclusion from content presented in the original text. To this end, we characterize elaborations by annotating their \emph{contextual specificity}, i.e., the extent to which the added content is specific to the current topic under discussion. 

We reveal that our dataset contains a fairly balanced distribution of contextual specificity. Qualitatively, while inserting definitions may help provide background about entities, highly contextualized elaborations interpreting or clarifying content can help readers understand the larger implications or significance of ideas presented in the original text. We propose the primary task of generating elaborations given document context. We present baselines for elaboration generation mainly using GPT-2~\cite{gpt2}, and discuss some of the challenges, especially with respect to the contextual specificity of added content.

We find that generation quality can be improved by selecting an elaboration with an appropriate predicted contextual specificity level. However, existing methods struggle to effectively incorporate input context to generate elaborations. We hope that this study will motivate advancement in elaborative simplification. 

In summary, our main contributions include: 
\begin{enumerate}
% \vspace{-0.3mm}
\item Introduction of elaborative simplification, a previously understudied phenomenon in text simplification;
% \vspace{-0.3mm}
\item A new, annotated dataset of 1.3K naturally occurring elaborations in the Newsela corpus and their contextual specificity;
% \vspace{-0.3mm}
\item Analysis of the challenges of elaborative simplification for pre-trained language models through performance of our baselines. 
\end{enumerate}

\noindent We make our annotations and code available at \url{https://github.com/nehasrikn/elaborative-simplification}. 

\section{Data and Annotation}
\begin{figure*}[h]
\centering
\includegraphics[width=\textwidth]{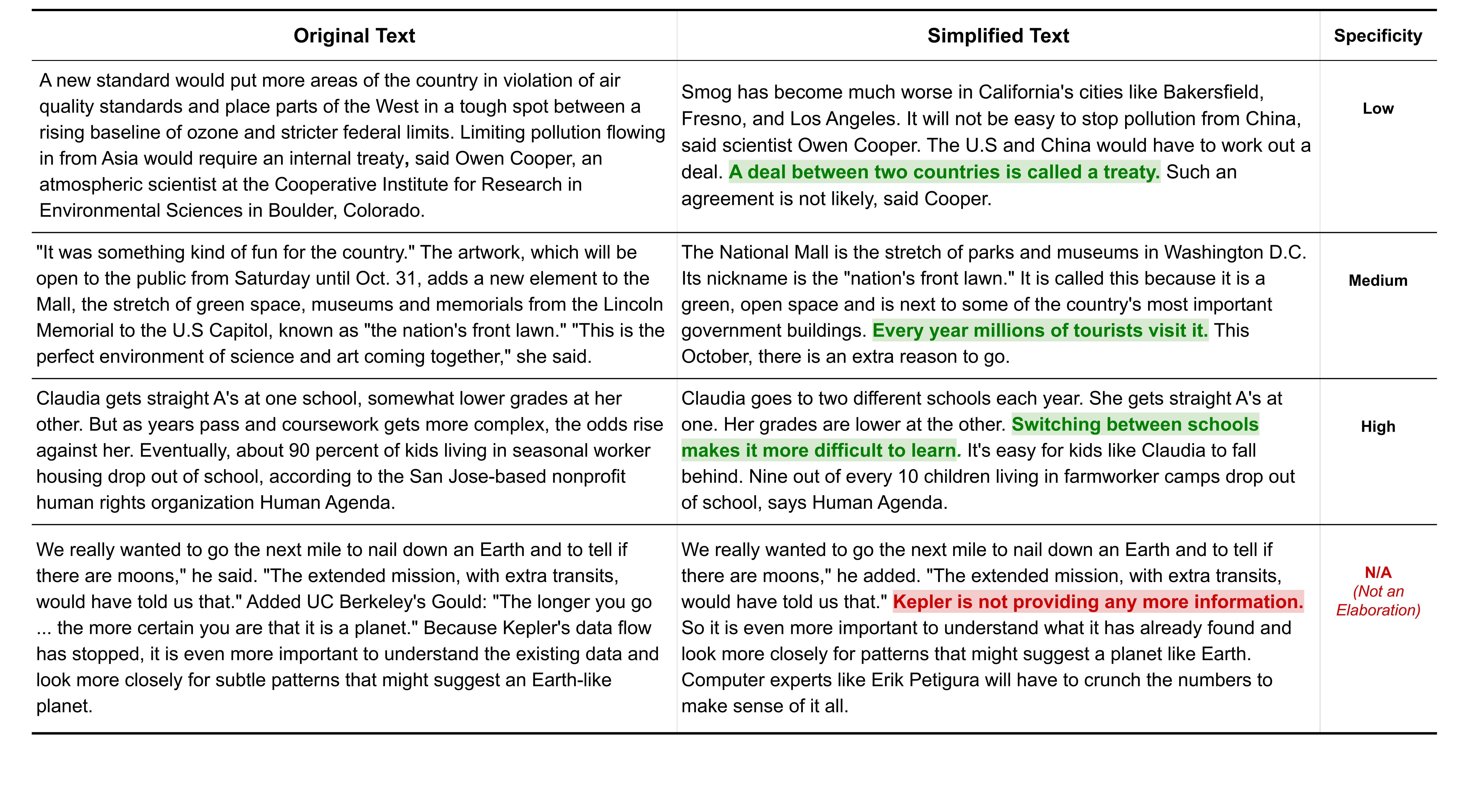}
\caption{Example candidate elaborations. Rows 1--3 contain verified elaborations. Row 4 contains a rejected candidate. We include the original and simplified text regions, highlighting the candidate elaboration, and its corresponding level of contextual specificity in Column 3.}
\label{fig:elaboration-examples}
\end{figure*}

Elaborative simplification involves the \emph{insertion} of content to make simplified text easier to understand. We present an annotated dataset of 1.3K elaborations from the Newsela corpus~\cite{xu2015problems}, which contains English news articles manually simplified by professional editors. We describe the scope of our elaborative simplification study (\S \ref{sec:scope}), strategies for trusted annotators to extract elaborations (\S \ref{sec:expert-pilot}) and rate contextual specificity (\S \ref{sec:data:contextualization}), and scaling up annotation through crowdsourcing with rigorous quality control (\S \ref{sec:crowd}).  

\subsection{What is an elaboration?}
\label{sec:scope}
We consider a sentence an elaboration if it contains new content (e.g. statements about entities, actions, or concepts) present in the simplified document, but semantically missing from the original document. Note that while elaborations can contain multiple sentences, we define our label at the sentence level. Past simplification research has focused on operations such as substitution and deletion, but simplifying a piece of text that may contain unknown or difficult concepts could involve inserting simple explanations as well. As we highlight in \S\ref{sec:related-work}, others have shown that audiences such as new language learners benefit from elaboration or explanation insertion (and conversely, that unfamiliar concepts negatively impact reading comprehension), though computational approaches till date have been largely limited to definition retrieval.

\paragraph{Scope.}
We intentionally choose to study \textit{how} concepts are elaborated, posing a scenario where an author has the freedom to specify where to elaborate, and our system generates an appropriate elaboration. We do this for two main reasons: first, understanding how to elaborate can be utilized in a system where users specify what to elaborate on, in the spirit of personalized simplification~\cite{paetzold2016anita,bingel2018lexi}. Second, determining \emph{when} to elaborate is arguably pragmatically more complex, in that the need for elaboration often relies on the writer's belief about their readers' background, knowledge, and reading ability, as well as their own judgments on how often to elaborate. For example, in the extreme case, inserting an elaboration after every sentence could prove useful for children or readers with no background knowledge about the document content, but may be unnecessary for adults or those with sufficient knowledge.

\paragraph{Task.} We introduce the primary task of elaboration generation: given some document context $C$ consisting of text from the original and/or simplified documents, generate an elaboration $E$. 

\subsection{Extracting Elaborations}
\label{sec:expert-pilot}

Detecting elaborative simplification requires crafting a way to reliably extract sentences containing new content in simplified documents. Asking humans to read and annotate every sentence in each document is prohibitively costly. To streamline this process, we first obtain candidate elaboration sentences with automatic sentence alignment, then use human annotation to extract true elaborations.

\paragraph{Candidate extraction.}
Each set of articles in the Newsela corpus consists of multiple simplified articles ranging from grades 3--12. We choose the article written for the lowest grade level as our simplified document (we leave investigating simplified documents across higher grade levels as future work). Using the approach from~\citet{jessy}, we then align sentences from the original and simplified documents by thresholding the cosine similarity of sentence vector representations using  Sent2Vec~\cite{pagliardini2018unsupervised}. 
We then consider sentences in the simplified document that are not aligned with any sentence in the original document as \emph{candidate elaborations}. 
Of the 54,892 sentences across the 1,042 simplified documents (on average, 52 sentences per document), 6,207 were extracted as candidate elaborations.

\paragraph{Human verification.}
Before crowdsourcing, we conducted a pilot study of elaboration verification with two sets of annotators: (1) Expert annotators (one graduate student, one undergraduate, both native speakers of English) who studied the data extensively; (2) 13 trusted undergraduate volunteer annotators at our university, also native English speakers. They received detailed instructions, but no face-to-face training. This allowed us to gauge task scalability and to gather feedback to design our crowdsourcing protocol. 
The 13 annotators each annotated a subset of 50 randomly selected documents (a total of 301 candidate elaborations) from our corpus. Each candidate elaboration was annotated by 2 to 4 annotators. 

For each original-simplified document pair, we provided annotators with the entirety of both documents. We asked them to identify whether each candidate elaboration truly contained semantically new content, and to provide a rationale for their annotation. We \emph{aggregated} the annotations for each candidate elaboration by taking the mode of all responses. The expert annotation consisted of 150 of these candidate elaborations under the same setup. Figure \ref{fig:elaboration-examples} shows some examples of verified and rejected candidate elaborations.

\paragraph{Agreement.} Cohen's Kappa among the two expert annotators is 0.75, indicating substantial agreement~\cite{artstein2008inter}. Cohen's Kappa between expert annotations and aggregated student annotations is also substantial, at 0.67. Krippendorff's alpha among the 13 student annotators is 0.37.
As in complex NLP annotations~\cite{nye2018corpus}, although there is subjectivity among individual annotators due to the complicated nature of the task, their aggregated judgment can be of as high quality as trained expert annotators.

\subsection{Contextual Specificity}\label{sec:data:contextualization}
At first glance, it seemed that elaborative simplification might simply involve retrieving definitions~\cite{paetzold2016anita} or crafting informative post modifiers~\cite{kang2019pomo}. However, while annotating candidate elaborations, we noticed that elaborations in our corpus took a variety of forms.

To better understand content addition, we conducted an extensive study of elaborations and found that often times, clarification or analysis sentences specific to document context are inserted to aid comprehension or facilitate connections between content in the original text. Notably, elaborations vary in their \emph{contextual specificity}, i.e., the degree to which an elaboration is specific to the context.\footnote{We draw a distinction between contextual specificity and contextual relevance (as in~\citet{kang2019pomo}).} 
For example, while simple definitions can be inserted into several different documents mentioning the same entity (low contextual specificity), some elaborations containing clarifications, commonsense reasoning applied to document content, or explicit inference are more contextually specific, as illustrated in Figure~\ref{fig:elaboration-examples}. 

This formulation is inspired by prior work in text specificity \cite{li2016improving,ko2019domain} which is related to how a sentence ``stands on its own" or sentence ``decontextualization" as in \citet{parikh2020totto, choi2021decontextualization}. As we discuss in \S \ref{sec:crowd}, contextually specific elaborations tend to have slightly lower sentence specificity, thus depending on the surrounding context to enhance understanding.
% \todo{\small All sentences will be similar unless in certain discourse relations. I think it's fair to just say that contextually specific sentences tend to be a bit more general, thus it depends more on the context.}

We ask the pair of experts from the previous pilot to annotate 116 randomly chosen verified elaborations for contextual specificity. Each expert was again given the entirety of the original and simplified documents with the highlighted elaboration, and asked to label its contextual specificity on a scale of 1--3 (low/medium/high). Their Fleiss' Kappa showed moderate agreement~\cite{landis1977measurement} with $\kappa=0.57$. Spearman's correlation between the two annotators is $0.72$. To enable collection, study, and modeling of this linguistic knowledge at scale, we gather contextual specificity ratings during crowdsourcing.

\subsection{Crowdsourcing}
\label{sec:crowd}

Annotating elaboration verification and contextual specificity requires careful reading and thoughtful reasoning over text. For the pilot described in \S \ref{sec:expert-pilot}, we provided thorough instructions and example documents and annotations. While these trusted annotators delivered high quality, reliable annotations, they ultimately cannot annotate a dataset of the scale supervised systems require. To remedy this, we use Amazon Mechanical Turk to collect labels at scale, albeit with slightly more noise. Our rationale is that models can tolerate this during training, and we ensure cleaner validation and test sets through expert annotations.

\paragraph{Task setup.} We ask workers to annotate elaboration verification and contextual specificity in a single task (HIT). For each candidate elaboration, we provide crowdworkers with the text region from the simplified document containing the elaboration, and the aligned text region from the original document.
We ask crowdworkers to categorize each candidate as a true elaboration, not an elaboration, or indicate that the snippets were unrelated. If true elaboration is selected for a candidate, we asked them to rate its contextual specificity\footnote{During crowdsourcing we utilized a 5-point scale, but aggregated the labels to a 3-point scale because the two scores on either end of the scale are not distinctive (i.e., are subjective).}. From feedback during our expert pilots, we determined that providing entire documents was often distracting, proving necessary only in rare cases where content was drastically rearranged. Instead, we display text regions of 5--7 sentences from both the simplified and original documents. The simplified text region contains the candidate elaboration and surrounding sentences, and the original text region contains sentences that are aligned with neighboring sentences of the elaboration in the simplified text region. We compose HITs that consist of $\sim$4 candidates from the same article.

\paragraph{Quality control.}
To ensure high quality annotations, we ask crowd workers to provide a rationale for each rating decision, as in \S\ref{sec:expert-pilot}. These rationales provide insight into worker interpretations of our task, allowing us to actively curate annotations to only include reliable annotations in our dataset. For example, using this method, we were able to remove annotations where crowd workers inflated specificity ratings due to coreferent entity mentions (i.e \textit{``It is a tube that moves blood"} as opposed to \hspace*{0.005em}\textit{``An artery is a tube that moves blood")}. 

In addition, we require all crowd workers to reside in the US, UK, Canada, Australia, or New Zealand, and to have completed $\geq$ 100 HITs with an acceptance rate of 95\%. Each elaboration is annotated by 5 different crowdworkers. Through active monitoring and small batches of HIT releases, we identified a set of workers that we trust and invite back to the task. Initially, we pay \$0.15 --\$0.23/HIT, and retroactively pay trusted workers at the rate of \$8/hr after work time information is obtained.

\paragraph{Agreement between trained and crowdsourced annotators.}
For both tasks, we aggregate crowdsourced labels by taking the mode of all responses\footnote{Using the mean as an aggregation function resulted in noisier labels.}. Cohen's Kappa of elaboration verification between crowdworkers and experts is 0.37 (fair). 
To measure contextual specificity agreement between crowdworkers and experts, we use Krippendorff's alpha with an ordinal distance metric, aggregating Turker and expert responses using the mode to obtain an agreement value of $\alpha = 0.47$, indicating moderate agreement~\cite{artstein2008inter}.
We attribute the disparity between inter-expert agreement and expert-crowdworker agreement to the challenge and subjectivity of this task, especially amongst untrained crowd workers. Though crowdsourcing our data does result in a slightly noisier training set, we are able to collect data for supervised learning and analysis at scale.

\begin{table}
\centering
\small
\begin{tabular}{lcccc} 
\toprule
 & Low & Medium & High & \textbf{Total}  \\
 \midrule
Train & $303$  & $349$  & $397$  & $\mathbf{1049}$  \\
Valid & $71$  & $39$ & $24$ & $\mathbf{134}$ \\
Test & $42$ & $34$  & $40$  & $\mathbf{116}$ \\ 
\midrule
Total & $406$  & $423$  & $470$  & $\mathbf{1299}$ \\
\bottomrule
\end{tabular}
\caption{Dataset distribution by contextual specificity.}
\label{tab:data-dist}
\vspace{-5mm}
\end{table}

\paragraph{Dataset analysis.} Using Mechanical Turk, we annotated 4,178 out of the 6,207 candidate elaborations from 1,042 documents. We obtained 1,299 verified elaborations, establishing an approximate 32\% conversion rate from candidate to verified elaborations. Note that since candidate elaborations are obtained automatically, this does not accurately reflect the true elaboration rate per document, but rather a lower bound. On average, the elaborations in are corpus are 7--13 tokens long. 

% \begin{figure}[]
%     \centering
%     \includegraphics[width=0.5\textwidth]{figures/token_len_distr.png}
%     \label{fig:tok-lens}
% \end{figure}

\begin{figure}[]
    \includegraphics[width=0.48\textwidth]{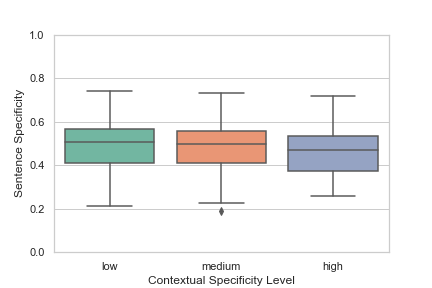}
    \caption{Sentence specificity distribution of elaborations across contextual specificity levels.}
\vspace{-5mm}
\label{fig:sent-spec}
\end{figure}

To ensure finetuning and evaluation quality, we use the expert-annotated subset of our data for the test set, and sought additional expert annotations for the validation set as well. Table \ref{tab:data-dist} shows our dataset size across splits, stratified by contextual specificity. Our dataset contains a relatively uniform distribution of specificity levels, confirming our qualitative analysis that the contextual specificity of added content is diverse.

\paragraph{Sentence Specificity.} As mentioned in \S\ref{sec:data:contextualization}, we explore the nature of sentence specificity of elaborations by running the sentence specificity predictor from~\citet{ko2019domain} on all standalone elaborations across all splits in our dataset. Sentence specificity predictions range on a continuous scale from 0 (very general) to 1 (highly detailed). Figure \ref{fig:sent-spec} shows the sentence specificity distribution across contextual specificity levels. The correlation between contextual and sentence specificity is $\tau=-0.11$, and is statistically significant. This negative correlation illustrates some of the intuition behind contextual specificity -- only when highly contextualized elaborations are inserted into documents do they facilitate document understanding.

%\footnote{Of the 54,892 sentences across the 1,042 simplified documents (on average, 52 sentences per document), 6,207 were extracted as candidate elaborations. Of these, we were able to annotate 4,178, and found 1,299 elaborations.}. 

\section{Elaboration Generation}
\label{sec:elab-gen}
We frame elaborative simplification as a natural language generation task, and describe a process mimicking editors elaborating as they compose a simplified document from the beginning (i.e. elaborations may be generated based only on the preceding simplified/original context) \footnote{We explored elaboration generation as a post-processing task after document simplification (Appendix~\ref{appendix:bart}). From preliminary results, we find it to be a more nuanced task which we leave for future work.}.
Elaboration generation is a challenging test for a model’s ability to produce relevant and effective elaborations ranging in contextual specificity given snippets of context from documents in our corpus. We investigate the abilities of pre-trained language models to generate elaborations, establishing baselines in \S\ref{sec:greedy-gen} and incorporating contextual specificity in \S\ref{sec:spec-gen}. We find that selecting elaborations of appropriate levels of predicted contextual specificity can help improve elaboration generation results.

\subsection{Baseline Elaboration Generation}  
\label{sec:greedy-gen}
We generate elaborations using GPT-2~\cite{gpt2}, a large-scale pre-trained language model which has been shown to be effective in a range of generation tasks, including in recent efforts to elicit world and commonsense knowledge ~\cite{zhou2020evaluating,yejin2020selftalk}.

Formally, we generate elaborations by conditioning on some document context, $C$. In this baseline setting, we generate sequences via greedy decoding. We utilize context from the original document ($C_o$) and from the simplified text ($C_s$). To understand the role that context plays in elaboration generation, we elicit elaborations from the language model by providing it one of the following: \textbf{(1)} 2 sentences prior to the gold elaboration in the simplified document (${C_{2s}}$), \textbf{(2)} a concatenation of 2 sentences prior to the gold elaboration from the simplified document and the corresponding aligned region in the original document (${C_{2s}+C_o}$), \textbf{(3)} 4 sentences prior to the gold elaboration in the simplified document (${C_{4s}}$).

\paragraph{Finetuning.} 
We finetune GPT-2 on the set of simplified documents written for the lowest grade level in the Newsela corpus, as well as on our dataset of verified elaborations excluding the test set. We found that such fine-tuning substantially improves generation quality (c.f.~Appendix~\ref{appendix:finetuning}).

\subsection{Specificity-guided Generation}  
\label{sec:spec-gen}
\begin{figure*}[]
    \centering
    \includegraphics[width=\textwidth]{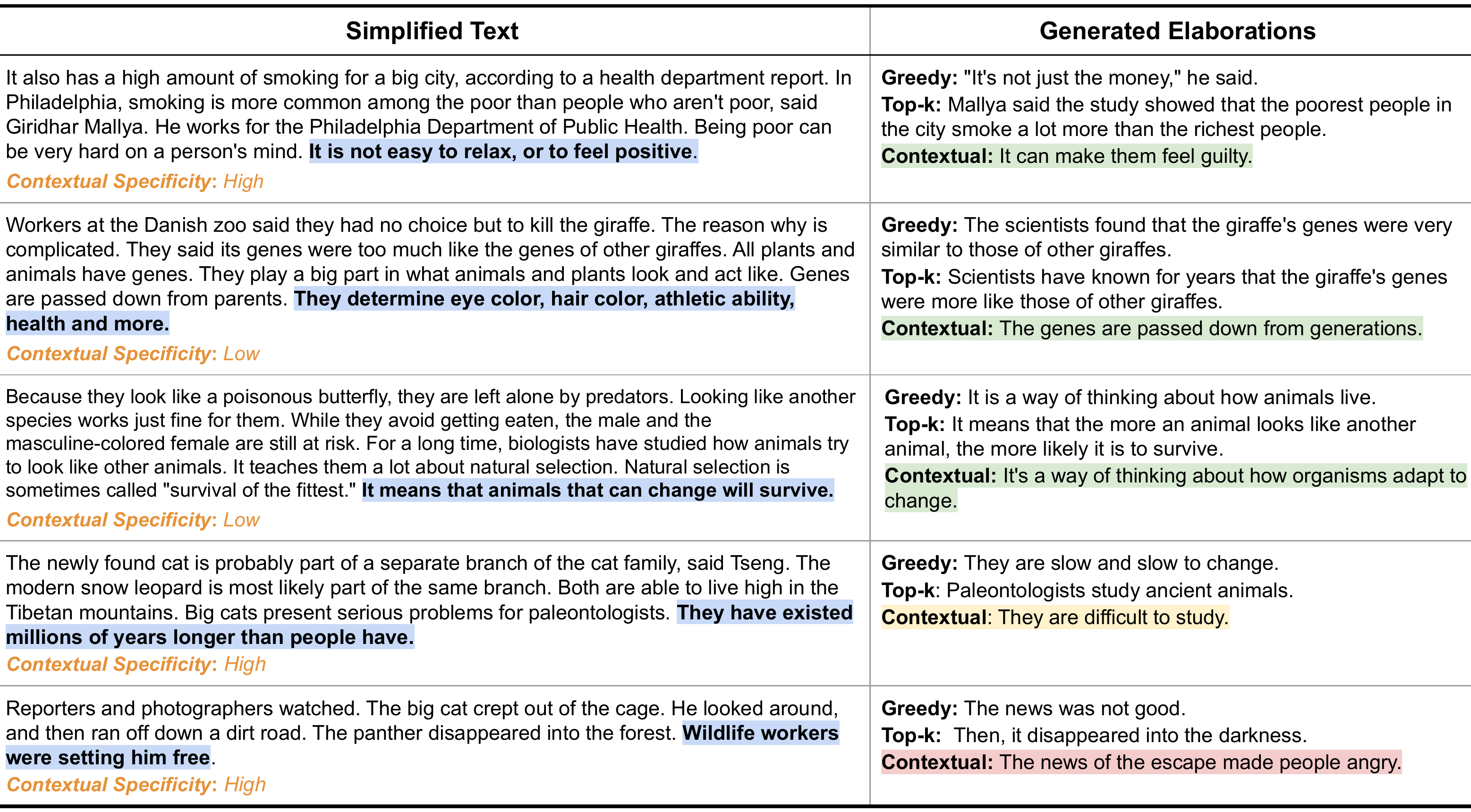}
    
    \caption{Examples of generated elaborations with the different decoding strategies described in \S \ref{sec:elab-gen}. Exs. 1--3 are cases where selecting a contextually-appropriate generated elaboration was effective. Ex. 4 is a relevant, sound elaboration with no content overlap with the gold elaboration, hence not rewarded by automatic metrics. Ex. 5 is a difficult case where context is essential -- the generated elaboration is not pertinent to document context.}
    \label{fig:gen-exs}
\end{figure*}

As discussed in \S\ref{sec:data:contextualization}, elaborations in our corpus are notably diverse in terms of their contextual specificity. Producing elaborations of appropriate contextual specificity is important, e.g., inserting an unnecessary definition instead of explaining a central concept can be ineffective or detrimental to readers' understanding. Rows 1-2 in Figure \ref{fig:gen-exs} show examples where the elaboration generated by the model in \S\ref{sec:greedy-gen} does not match the level of contextual specificity of the gold elaboration, motivating our exploration of including contextual specificity and its prediction to aid elaboration generation.

\paragraph{Contextual specificity prediction.}
We build a model to classify the level of contextual specificity of an elaboration as low, medium, or high to incorporate downstream during generation.
%\footnote{We have experimented with ordinal regression methods but classification yielded much better results.} 
We leverage BERT~\cite{bert} for this task. Appendix~\ref{appendix:contextual-specificity} explores this auxiliary task further to understand modern NLP models' ability to capture this linguistic information.

We train the model on $(E, s)$ pairs, where $E$ is an elaboration, and $s$ is its labeled contextual specificity. We feed $E$ as input to BERT, and then feed the \texttt{[CLS]} token embedding into an output layer for classification. We freeze the BERT parameters since fine-tuning yielded unstable results. We utilize \texttt{bert-base} from the HuggingFace library ~\cite{wolf2019huggingface}. After tuning on the validation set, we train for 5 epochs, using a batch size of 32 and a learning rate of 2e-3. We use the default dropout rate of 0.1 for self-attention layers, but refrain from adding dropout on our linear layer.

This contextual specificity model achieved an accuracy of $56.8 \pm 1.5$, a macro-averaged F1 score of $55.3 \pm 1.6$, a Spearman correlation of $47.5 \pm 2.6$, and a mean absolute error of $0.552 \pm 0.01$, averaged across 15 randomly initialized runs. This performance is better or on par with other models that incorporate document context in different ways (Appendix~\ref{appendix:contextual-specificity}). We find contextual specificity prediction to be a challenging task for BERT. Prediction of \emph{expected} contextual specificity (i.e prediction from context alone, without the elaboration) was particularly difficult, and we leave building stronger models in this setting to future work.

\paragraph{Generation.}
We investigate the importance of contextual specificity in generating effective elaborations by comparing sequences generated in 3 ways: 
\begin{packed_enumerate}
\item \textbf{Greedy:} Generate elaborations via greedy decoding. This setting was discussed in \S \ref{sec:greedy-gen}.
\item \textbf{Top-k:} Sample a sequence from the language model using top-k sampling~\cite{fan2018hierarchical}, without considering contextual specificity.
\item \textbf{Contextual specificity-informed sampling}, shorthand \textbf{Contextual:} Sample sequences using top-k sampling until we have 3 elaborations of low, medium, and high contextual specificity, as predicted by the contextual specificity model, and select the sequence with predicted contextual specificity matching the gold specificity level.
\end{packed_enumerate}

In practice, one would ideally use a contextual specificity model trained \emph{without} the elaboration itself (i.e., \emph{Context-Only} models in Appendix~\ref{appendix:contextual-specificity}) to predict the appropriate level of contextual specificity of a generated elaboration. However, since we leave to future work to build a strong model presented with this setup, we instead utilize the gold specificity label and explore the upper bound with our generation experiments.

We use sampling-based decoding strategies to achieve contextual specificity diversity because we find that while beam-based decoding methods may result in sequences with diverse \textit{content}, they do not necessarily result in sequences with diverse \textit{contextual specificity.}

\subsection{Experimental Settings} 
We use GPT-2 medium from the HuggingFace library ~\cite{wolf2019huggingface} to finetune and generate elaborations. We finetune GPT-2 on documents simplified for the lowest-grade level in the Newsela corpus for 3 epochs with a learning rate of 1e-5 and a batch size of 32. For sampled sequences, we use top-k sampling with $k=40$, and a temperature of $t=0.45$, tuned on validation data.

\section{Generation Evaluation}
\label{sec:evaluation}
As elaboration generation is a new task, we include BLEU scores for completeness and emphasize human evaluation, which provides important insight early on in the study of a new phenomenon.

\subsection{Automatic Evaluation}
\label{sec:automatic-eval}

We report BLEU~\cite{papineni2002bleu}, a standard metric in generation tasks. Table~\ref{tab:generation-results} shows corpus BLEU-1 and BLEU-2 scores on our test set.
As illustrated in Table~\ref{tab:generation-results}, the best models, as reflected by BLEU, are those finetuned on the Newsela simplified corpus, with four sentences from the simplified document before the gold elaboration as context.

While BLEU captures lexical overlap between generated and gold elaborations, it is also criticized due to poor correlation with human judgments~\cite{liu2016not,novikova2017we,chaganty2018price}, as it fails to capture semantic similarity or reward multiple plausible hypotheses. During manual inspection of these sequences, we find that elaborations produced after finetuning GPT-2 can be semantically plausible, coherent, and elaboration-like. Content that is pertinent and new, but that does not overlap with the content in the gold elaboration is not rewarded. In some cases, staying true to the content of the gold elaboration is likely unnecessary, as long as the contextual specificity is comparable (see row 4 in Figure~\ref{fig:gen-exs}). To that end, we also perform a human evaluation study of generated elaborations, given that the purpose of elaborations is largely to make simplified text easier to understand for readers.

\subsection{Human Evaluation}

\begin{table}
\setlength{\tabcolsep}{5pt}
\small
\begin{tabular}{lcc|cc|cc} 
\toprule
\multicolumn{1}{c}{} & \multicolumn{2}{c|}{Greedy} & \multicolumn{2}{c|}{Top-k} & \multicolumn{2}{c}{Contextual} \\ 
\midrule
Context & B-1 & B-2 & B-1 & B-2 & B-1 & B-2 \\ 
\midrule
$C_{2s}$ & $20.8$ &$6.77$  & $20.4$  & $6.12$  & $21.4$  & $7.26$  \\
$C_{2s}+C_o$  & $18.7$  & $5.66$  & $17.2$  & $4.32$  & $19.0$  & $5.31$  \\
$C_{4s}$  & $20.8$  & $5.54$  & $19.7$  & $6.06$  & $\mathbf{22.4}$  & $\mathbf{7.56}$  \\
\bottomrule
\end{tabular}
\centering
\caption{\label{tab:generation-results}BLEU-1 and BLEU-2 scores for elaborations generated by GPT-2, finetuned on the Newsela simplified document corpus. Results for our best model, which we conduct human evaluation on, are in bold.}
\end{table}

\begin{table}[]
    \centering
    \small
    \begin{tabular}{c|ccc}
    \toprule
        System & Greedy & Top-k & Contextual \\ \midrule
        \% selected & $53.2$ & $44.9$ & $58.0$ \\
    \bottomrule
    \end{tabular}
    \caption{Percentage of annotations for which users selected elaborations generated by each model.}
    \label{tab:generation-humaneval}
\end{table}

We set up our human evaluation similar to~\citet{panthaplackel2020learning}, providing a pair of expert evaluators elaborations generated by our $C_{4s}$ model (see Table~\ref{tab:generation-results}) in each of the three setups (greedy, top-k, contextual), and ask them to select the sequence they thought was most coherent, topical, semantically plausible, and elaboration-like. We allow selection of multiple sequences if they are equally good, and no selection if all sequences are poor. We report human evaluation results as the percentage for which evaluators chose the sequence as higher quality. Two annotators each annotated all 116 examples in our test set, resulting in 232 evaluations total. Table~\ref{tab:generation-humaneval} shows these results. We calculate human agreement via Cohen's kappa with MASI distance~\citep{passonneau2006masi}, obtaining $\kappa=0.51$, indicating moderate agreement~\citep{artstein2008inter}. This round of evaluation confirmed that incorporating contextual specificity is helpful, consistent with our findings with BLEU.

\section{Analysis and Discussion}
\label{sec:discussion}
We observe that GPT-2, finetuned on simplified text from the Newsela corpus, is able to adopt elaborative style (i.e short sentences of 7--13 tokens with limited vocabulary), see Figure \ref{fig:gen-exs}. We find that the model can be effective at generating simple definitions and reasoning. However, the content contained in the elaborations is often not anchored in the document itself -- generated sequences seem relevant to the snippet of context provided, but less so when placed in the larger document (see row 5 of Figure \ref{fig:gen-exs}).

\paragraph{Original Text.} We observe that our best model involves context only from the simplified document. We attribute the drop in performance of models with $C_o$ as a part of input largely to the crude incorporation of content from the original document, which is stylistically starkly different from simplified text, most notably in terms of length and vocabulary complexity. Since one of the main sources of relevant content during simplification is the original document, better methods to incorporate text or information from the original document is an important direction for future work.

\begin{figure}[t]
\centering
\includegraphics[width=0.46\textwidth]{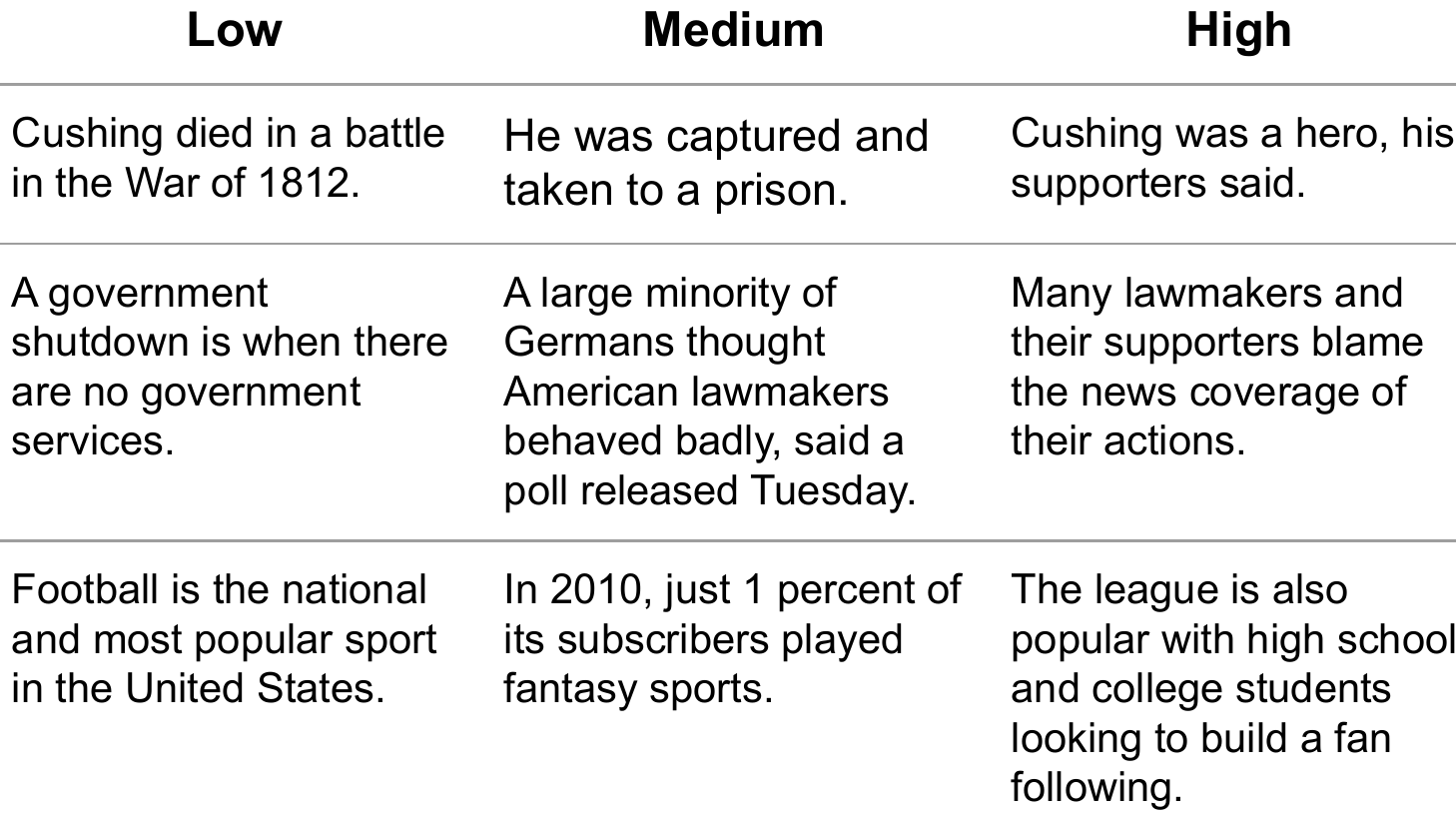}
\caption{Example generated elaborations of varying contextual specificity.}
\label{fig:gen-varying-cs-exs}
\vspace{-1.3mm}
\end{figure}

\paragraph{Effectiveness of contextual specificity.} Decoding with top-k sampling allowed GPT-2 to generate low, medium, and high contextualization sequences. A few examples of generated elaborations with varying contextual specificity that were conditioned on the same context are shown in Figure~\ref{fig:gen-varying-cs-exs}. 
For most of our models, we do see an improvement when appropriately contextually specific sequences are chosen (rows 1--3 in Figure \ref{fig:gen-exs}), suggesting the importance and need for further improvement of contextual specificity models.

While our methods take contextual specificity into account, they do not consider factuality or larger document relevance. An improved decoding scheme considering these could promote sequences that better align with larger document context.

\paragraph{Retrieval.} Elaborations of medium to high contextual specificity often involve external knowledge not readily available from the simplified or original text. For example, generating factually correct details about a certain event or entity with little to no background on the event the document is referring to can prove challenging for pre-trained language models. To that end, generating truly effective elaborations of medium to high contextual specificity may require some type of retrieval module.

\section{Related Work}
\label{sec:related-work}
Text simplification has been studied extensively~\cite{siddharthan2014survey}, especially at the sentence level. Recent progress has largely been driven by adapting monolingual translation for sentence simplification~\cite{wubben2012sentence,wang2016text,xu2016optimizing,zhang2017sentence,dong2019editnts,kriz2019complexity}. This paradigm, while effective at transforming text, does not suffice when \textit{new} content needs to be generated. A recent survey~\cite{alva2020data} identifies explanation generation in simplification as an understudied area in dire need of new resources and methods. We tackle content addition, framed as explanation generation during simplification, and name it broadly as \emph{elaborative simplification}.

The need for elaborative simplification is highlighted in prior hand-coded analysis~\cite{yano1994effects}, which showed that language learners and other audiences benefit from insertion of relevant elaborations and explanations, and that new or unfamiliar concepts negatively impact reading comprehension~\cite{kintsch1985reading}. However, existing computational approaches are limited to the retrieval of definitions~\cite{damay2006simtext,kandula2010semantic,eom2012sense,paetzold2016anita}, or constrained tasks such as post-modifier generation~\cite{kang2019pomo}.

\section{Conclusion}
We presented the first data-driven study of elaborative simplification, i.e., content insertion during text simplification. We constructed a new corpus of 1.3K verified elaborations, observing a spectrum of contextual specificity and rich types of added content. We developed baselines for elaboration generation using pre-trained language models and found that considering contextual specificity could improve generation quality. We discussed some of the challenges of generating elaborations, and call for techniques to address elaborative simplification.

\section*{Acknowledgments}
We thank Greg Durrett for reviewing an early draft of this paper, and Joel Tetreault and the UT Austin Computational Linguistics group for valuable feedback and discussions. We also thank all the student and crowd annotators that contributed to this dataset. This work was partially supported by NSF Grant IIS-1850153. We also acknowledge the Texas Advanced Computing Center (TACC) at UT Austin for providing the computational resources for many of the results within this paper.

\bibliographystyle{acl_natbib}
\bibliography{anthology,acl2021}

\clearpage

\appendix
\begin{table*}[ht!]
\centering
\small
\begin{tabular}{llcccc} 
\toprule
 & Context & Acc. & F1 & Correlation & MAE \\ 
\midrule
 & $C_o + C_{4s}$ & $45.2 \pm 3.0$  & $43.1 \pm 2.8$  & $27.8 \pm 4.9$  & $0.729 \pm 0.05$  \\
\textit{Context Only}  & $C_{4s}$ & $46.4 \pm 2.9$  & $44.9 \pm 3.0$  & $32.4 \pm 4.4$  & $0.679 \pm 0.04$  \\
 & $C_o$ & $37.9 \pm 4.6$  & $36.1 \pm 5.4$  & $20.2 \pm 1.0$  & $0.813 \pm 0.07$  \\ 
\midrule
 & $E$ & $\mathbf{56.8 \pm 1.5}$  & $\mathbf{55.3 \pm 1.6}$  & $47.5 \pm 2.6$  & $0.552 \pm 0.01$  \\
\textit{With Elaboration}  & $C_o + C_{4s} + E$ & $50.5 \pm 3.8$  & $48.3 \pm 4.0$  & $40.4 \pm 5.8$  & $0.628 \pm 0.05$  \\
 & $C_{4s} + E$ & $55.3 \pm 3.3$  & $54.0 \pm 2.5$  & $\mathbf{50.8 \pm 4.1}$  & $\mathbf{0.545 \pm 0.03}$  \\
 & $C_o + E$ & $43.7 \pm 1.8$  & $41.7 \pm 2.0$  & $26.7 \pm 3.6$  & $0.749 \pm 0.03$  \\
\bottomrule
\end{tabular}
\caption{\label{appendix:tab:specificity-perf} Contextual Specificity Prediction results, including accuracy, macro-averaged F1, Spearman's correlation, and Mean Absolute Error, reported across 15 runs. We bold our best results. The performance differences between (1) $C_{4s} + E$ vs $E$, (2) $C_o + C_{4s}$  vs $C_{4s}$, and (3) $C_o + C_{4s} + E$  vs. $C_{4s} + E$ are not statistically significant.}
\end{table*}

\section{Contextual Specificity Prediction}
\label{appendix:contextual-specificity}
We further explore the auxiliary task of contextual specificity prediction introduced in \S \ref{sec:spec-gen}, prompted by the observation of diverse elaborations in our corpus. Formally, the task involves predicting the contextual specificity $s$ of an elaboration $E$ as low, medium, or high, given some document context $C$.

\subsection{Methods}
As described in \S\ref{sec:spec-gen}, we use BERT~\cite{bert} for this classification task. We do so in two settings based on surrounding text and/or the actual elaboration. Settings which include the elaboration can aid generation models by utilizing generated hypothesis elaborations and surrounding text to select sequences that are appropriately contextually specific. Settings that operate off context alone capture the \emph{expected} level of specificity. In addition to the $E$-only model presented in \S\ref{sec:spec-gen}, we explore combinations of $E$, $C_o$ (original document context) and $C_{4s}$ (4 sentences prior to the gold elaboration from the simplified document).

\paragraph{With elaboration.} We feed the input sequence into BERT and use \texttt{[CLS]} token representation of the sequence, projecting it using a weight matrix $W \in \mathbb{R}^{d x 3}$. Input sequences with the elaboration consist of \texttt{[CLS] $C$ [SEP] $E$}, where $C$ is either $C_o$ or $C_{4s}$, or both. When both types of context are used, we learn a representation for a separation token \texttt{[CONTEXT\_SEP]} to distinguish between the two, and use $C = C_o \texttt{[CONTEXT\_SEP]} C_{4s}$. 

\paragraph{Context only.} While contextual specificity clearly involves the elaboration itself, context-only models help us understand whether it is predictable from context alone, and simulate a realistic setting during simplification, when these models may be incorporated before the actual elaborative text is generated. Input to these models is crafted similarly, but excluding $E$ from the sequence.

\subsection{Experiments and Analysis} 
We train on $(E,s)$ pairs, and utilize \texttt{bert-base} from the HuggingFace Transformers library. We feed the sequence representation from the \texttt{[CLS]} token embedding into an output layer for classification \footnote{We tried finetuning our contextual specificity prediction models on our elaboration dataset, but found that our dataset was too small to yield stable results.}. For each setting, we train for 5 epochs, using a batch size of 32, and a learning rate of 2e-3. We use the default dropout rate of 0.1 for self-attention layers, but refrain from adding dropout on our linear layer.

\vspace{0.3em}
\noindent\textbf{Results.} We use the same four metrics to evaluate our results -- two classification metrics (accuracy, macro-averaged F1), and two regression metrics (Spearman's correlation and mean absolute error), and we again report mean performance over 15 different, randomly initialized runs. Results are shown in Table ~\ref{appendix:tab:specificity-perf}, and suggest that this is a challenging task, even for powerful pre-trained language models. The best predictor of contextual specificity, in terms of correlation and MAE, is context in the form of 4 sentences before the elaboration combined with the elaboration itself. However, the elaboration-only model performs the best in terms of accuracy and F1.

\vspace{0.3em}
\noindent\textbf{Original Text Presence.}  In all settings in which the aligned snippet of text from the original document was fed in as partial or complete input to the model, we see a reduction in performance. Compared to text from the simplified document, text from the original document is stylistically distinct. Consequently, when jointly fed in as context with simplified text, the input is largely incoherent, potentially impacting the model. We leave studying more effective ways of incorporating context from the original document to future work.

\vspace{0.3em}
\noindent\textbf{Qualitative Analysis.}
In cases where linguistic cues explicitly indicate the level of contextual specificity, our model performs well---i.e when definitions are inserted as \textit{"A is B"} or reasoning is inserted as \textit{"A but B"} or \textit{"The reason for A is B"}. However, predicting the contextual specificity of more nuanced sentences may require an improved method of modeling surrounding context. For example, when the elaboration contains a definition of a term from a different sentence using coreferent mentions, our model predicts a higher level of contextual specificity. 
In general, our model over-predicts highly contextualized elaborations, and under-predicts lower levels of contextual specificity. Medium contextual specificity was hardest for our models to predict accurately.

\vspace{0.3em}
\noindent\textbf{Amount of context.} To understand the impact of the amount of context on performance, we vary the number of sentences ($\{2,4,6\}$) before the elaboration to feed into our best performing model involving context ($C_s + E$). Table \ref{tab:simple-context} shows these results. We see that merely increasing the amount of context fed to the model does not translate to stronger results -- considering overall performance, 4 sentences before the elaboration from the simplified document performed best.

\begin{table}[t]
\setlength{\tabcolsep}{2pt}
\centering
\small
\begin{tabular}{l|l|l|l|l} 
\toprule
\multicolumn{1}{l}{} & \multicolumn{1}{c}{Acc.} & \multicolumn{1}{c}{F1} & \multicolumn{1}{c}{Corr} & \multicolumn{1}{c}{MAE} \\ 
\midrule
$C_{2s}$  & $53.6\pm1.8$  & $51.2\pm3.1$  & $\mathbf{51.7\pm6.8}$  & $0.566\pm0.05$  \\
$C_{4s}$  & $\mathbf{55.3\pm3.3}$  & $\mathbf{54.0\pm2.5}$  & $50.8\pm4.1$  & $\mathbf{0.545\pm0.03}$  \\
$C_{6s}$  & $53.9\pm4.0$  & $52.0\pm3.6$  & $44.3\pm5.5$  & $0.591\pm0.04$  \\
\bottomrule
\end{tabular}
\caption{Mean performance of $C_s + E$ model over 15 runs with varying amounts of pre-elaboration context.}
\label{tab:simple-context}
\end{table}

\section{Elaboration Generation}

\subsection{GPT-2 Finetuning}
\label{appendix:finetuning}
We explore generation with GPT-2 across varying finetuning settings -- (1) zero shot (no finetuning, only relying on GPT-2's pre-training), (2) finetuning on the set of simplified documents in the Newsela corpus (excluding documents from the test set), and (3) finally on our elaboration corpus. We utilize the same 3 decoding schemes described in \S~\ref{sec:spec-gen} across these different finetuning settings. We used a temperature of $t=0.7$ for the zero shot setting, and $t=0.45$ for finetuned settings. For finetuning on our elaboration corpus, we trained for 3 epochs with a batch size of 8 and a learning rate of 1e-3. We report BLEU-1 and BLEU-2 as described in \S ~\ref{sec:automatic-eval}. As BLEU metrics for setting 2 are already included in Table~\ref{tab:generation-results}, we report metrics for zero-shot generation (Table~\ref{appendix:tab:zero-shot-results}), and for generation after finetuning on our elaboration corpus (Table ~\ref{appendix:tab:task-finetuning}). Comparatively, finetuning GPT-2 on the set of simplified Newsela documents yielded the best performance, and we attribute this to there being strictly more data in that setting as opposed to our corpus of verified elaborations.

\begin{table}[h]
\setlength{\tabcolsep}{5pt}
\small
\begin{tabular}{lcc|cc|cc} 
\toprule
\multicolumn{7}{c}{\textbf{Pre-trained GPT-2}} \\
\multicolumn{1}{c}{} & \multicolumn{2}{c|}{Greedy} & \multicolumn{2}{c|}{Top-k} & \multicolumn{2}{c}{Contextual} \\ 
\midrule
Context & B-1 & B-2 & B-1 & B-2 & B-1 & B-2 \\ 
\midrule
$C_{2s}$  & $12.48$ & $2.71$ & $9.82$~ & $2.04$ & $11.93$ & $2.66$ \\
$C_{2s}+C_o$  & $12.21$  & $2.58$  & $9.80$  & $2.08$  & $10.86$ & $2.82$ \\
$C_{4s}$  & $13.46$ & $3.35$ & $11.78$ & $2.43$ & $13.80$  & $3.89$  \\
\bottomrule
\end{tabular}
\centering
\caption{BLEU-1 and BLEU-2 for generation after finetuning on our elaboration corpus.}
\label{appendix:tab:zero-shot-results}
\end{table}

\begin{table}[h]
\setlength{\tabcolsep}{5pt}
\small
\begin{tabular}{lcc|cc|cc} 
\toprule
\multicolumn{7}{c}{\textbf{Fine-tuned GPT-2: Elaboration Corpus}} \\
\multicolumn{1}{c}{} & \multicolumn{2}{c|}{Greedy} & \multicolumn{2}{c|}{Top-k} & \multicolumn{2}{c}{Contextual} \\ 
\midrule
Context & B-1 & B-2 & B-1 & B-2 & B-1 & B-2 \\ 
\midrule
$C_{2s}$  & $20.9$ & $6.82$ & $19.11$~ & $5.32$ & $19.38$ & $5.47$ \\
$C_{2s}+C_o$  & $11.89$  & $2.78$  & $12.72$  & $2.77$  & $14.2$ & $3.05$ \\
$C_{4s}$  & $20.17$ & $5.87$ & $16.89$ & $4.09$ & $18.97$  & $5.16$  \\
\bottomrule
\end{tabular}
\centering
\caption{BLEU-1 and BLEU-2 for the zero-shot generation setting.}
\label{appendix:tab:task-finetuning}
\end{table}

\begin{table}[ht]
\centering
\small
\begin{tabular}{lllll} 
\toprule
 & $C_{2s}$ & $C_{2s+}$ & $C_{4s}$ & $C_{4s+}$ \\ 
\midrule
B-1 & $18.9$ & $21.5$ & $20.2$ & $20.1$ \\
B-2 & $5.05$ & $6.68$ & $6.02$ & $6.18$ \\
\arrayrulecolor{black}
\bottomrule
\end{tabular}
\caption{\label{appendix:tab:bart-results}
BLEU-1 and BLEU-2 for greedy generation with BART.}
\end{table}

\subsection{Generation with BART}
\label{appendix:bart}
In addition to GPT-2, we experimented with BART~\cite{bart}, a pre-trained sequence to sequence model. The encoder-decoder nature of BART allows us to explore elaborative simplification as a post-processing/post-editing scenario, where the model can receive context both preceding \textit{and} following the elaboration in the simplified text.

We finetune \texttt{bart-base} available via the HuggingFace Transformers library, and feed in four different types of context (1) $C_{2s}$, (2) $C_{4s}$, (3) $C_{2s+}$, (4) $C_{4s+}$. The latter two context settings utilize two and four sentences before and after the elaboration (without the elaboration itself). In all settings, the gold elaboration was the target. We finetune for 3 epochs, with a batch size of 2, and a learning rate of 1e-4, and generate elaborations via greedy decoding. Results are shown in Table~\ref{appendix:tab:bart-results}.

We find that BART is able to adopt elaborative style, generating short sequences with limited vocabulary, however we observe that the smaller size of our corpus affected BART's ability to generate coherent, diverse elaborations. In addition, we note that framing elaborative simplification as a post-processing task is a more difficult, nuanced setting -- the generated elaboration to be inserted must maintain the flow of the text and blend with the content present subsequent sentences. Elaborative simplification in this setting is another interesting, rich direction for future work.

\end{document}